\documentclass[11pt]{article}
\usepackage[preprint]{acl}
\usepackage{times}
\usepackage{latexsym}
\usepackage[T1]{fontenc}
\usepackage[utf8]{inputenc}
\usepackage{microtype}
\usepackage{inconsolata}
\usepackage{graphicx}
\usepackage{booktabs}
\usepackage{amsmath}
\usepackage{multirow}
\usepackage{array}
\newcolumntype{R}[1]{>{\raggedright\arraybackslash}p{#1}}

\title{Planning or Improvisation? Stress-Testing the Poetry Planning Site\\ on Open Models and Open Cross-Layer Transcoders}

\author{\'{E}ric Jacopin \\
  Cosmic AI, France \\
  \texttt{eric.jacopin@protonmail.com}}

\begin{document}
\maketitle

\begin{abstract}
\citet{lindsey2025biology} report that Claude 3.5 Haiku plans rhymes: features for candidate rhyme words are active on the newline before a line is written, and a suppress-and-inject intervention redirects the line only when applied there (their Figure~13). We test how far this generalizes on seven cells crossing four open models (0.6B to 2.6B parameters) with six open cross-layer transcoders (CLTs), on one consumer GPU, decomposing the claim into position specificity (C1), newline site identity (C2), and a newline-resident plan (C3). This is a stress test rather than a faithful reproduction: attribution graphs are unavailable for these CLTs, so features are found bottom-up from decoder vectors. C1 generalizes, in every cell and in all 247 of 444 prompt-by-inject pairs with a detectable effect, but the effective position is the final prompt token, adjacent to emission, and only two cells reach behaviorally meaningful probabilities. C2 and C3 are not recovered by any probe: a census of every active feature finds no rhyme-anticipating enrichment at the newline, and steering the newline while the model composes the whole line, over 36 runs and 8{,}640 sampled lines, shows why. That intervention is strong but one token long, making the injected word the first word of the composed line in 703 of 720 samples and leaving the rhyme six words later untouched. A final test drops the transcoder entirely: patching the newline's whole residual, at every layer, from a minimal-pair poem whose third line ends on a different rhyme moves the rhyme in 11 of 1{,}260 composed lines against 4 at baseline, with a design resolving 1.4\%. We read this as a boundary condition rather than a refutation: at this scale and with these transcoders, the causal site is emission-adjacent. We reproduce Figure~13's shape, not its mechanism. Code and data are public \citep{paperrepo2026}.
\end{abstract}

\section{Introduction}

Whether language models plan ahead is a central question for interpretability, and rhyming poetry is its cleanest published test case. \citet{lindsey2025biology} distinguish two mechanisms by which a model could end a line with a rhyme: \emph{pure improvisation}, choosing the word only when it must be emitted, and \emph{planning}, committing to a candidate ending before the line is written and composing toward it. For Claude 3.5 Haiku they report evidence for planning, and the sharpest piece of that evidence is a steering-location experiment (their Figure~13): suppressing the planned-word features and injecting an alternative changes the completion \emph{only} when the intervention is applied at the newline token preceding the line, several tokens upstream of the emitted word.

Figure~13 is an attractive target for a generalizability study. It is causal rather than correlational, it makes a falsifiable point prediction (effect concentrated at one specific, upstream position), and it was produced with a proprietary model and a proprietary 30M-feature cross-layer transcoder (CLT), so it has not been checked end to end outside Anthropic. Recent open-model work approaches it from different angles: \citet{hanna2026latent} reproduce rhyme planning on Qwen3 models with CLTs and argue that latent planning emerges with scale, and \citet{maar2026plan} shift rhyming behavior with contrastive steering vectors. \citet{jacopin2026prolepsis} reports position-specific CLT steering on Gemma 2 2B and Llama 3.2 1B, but at a site that differs from Anthropic's, a discrepancy that prior work has not analyzed and that we make explicit and central here.

\paragraph{What this paper is and is not.}
This is a stress test of the generalizability of Figure~13, in the sense of the reproducibility track it was written for: the same intervention machinery, applied beyond the original's experimental scope (open models two to three orders of magnitude smaller than Claude 3.5 Haiku, open CLTs ten to two thousand times narrower). It is \emph{not} a faithful reproduction of the original method. The original discovers planned-word features top-down, by attribution from a specific completion; no attribution-graph tooling exists for these open CLTs at consumer scale, so we discover features bottom-up, from their decoder vectors. This difference is load-bearing and we return to it repeatedly: bottom-up discovery finds features that \emph{write} a rhyme word, and can only test whether a plan is visible through such features. We state, for each claim, what our probes can and cannot show (\S\ref{sec:original}, \S\ref{sec:discussion}).

\paragraph{Contributions.}
Following the track's themes, we test \textbf{generalizability} (across open models and CLTs), \textbf{ablation} (which protocol components carry the effect), \textbf{baselines} (non-CLT steering, per-layer transcoders, and the random controls of the dead-salmon lineage, \citealp{meloux2025dead}), and \textbf{evaluation} (which metric choices change the conclusion). Our answers, from seven (model, CLT) cells swept over intervention position and strength, plus a re-analysis of 444 prompt-by-inject pairs:

\noindent\textbf{1. Position specificity (C1) generalizes, but is partly generic.} In 7/7 cells the intervention is effective at exactly one position, and in 247/247 pairs with a detectable effect that position is the final prompt token (\S\ref{sec:localization}). Random write-directions also concentrate their leverage there, at far smaller magnitudes: the \emph{location} of the spike is a property of the emission-adjacent write path, its \emph{magnitude} is what distinguishes the rhyme features (\S\ref{sec:generic}). \textbf{2. Behavioral redirection in two of seven cells.} Only the two mntss cells reach probabilities that change generated text (0.48, 0.85); the five Qwen3 cells move probabilities of $10^{-7}$ to $10^{-3}$, detectable but not behavioral (\S\ref{sec:localization}). \textbf{3. The newline site (C2) and a newline plan (C3) are not recovered.} The newline is inert for the rhyme target in every cell and every pair and carries no enrichment of rhyme-anticipating features (\S\ref{sec:site}). Steered while the model composes the whole line, it captures the next token and nothing beyond it: the injected word opens the line in 1{,}540 of the 1{,}545 lines it appears in, while the rhyme is never redirected under any of six counting criteria (\S\ref{sec:horizon}). \textbf{4. Magnitude belongs to the transcoder, not the model} (six orders of magnitude across CLTs, \S\ref{sec:magnitude}). \textbf{5. The protocol's free parameters matter:} the original documents no steering strength; the optimum varies from $s{=}1$ to $s{=}25$ across cells, and swapping one inject feature moves a cell from $3.5\times$ to $33{,}860\times$ (\S\ref{sec:ablations}).

All experiments run on one consumer GPU\footnote{An RTX 5060 Ti (16\,GB), Rust/candle stack \citep{candle2023,candlemi2026}. Commands, artifacts, runtimes, and feature identifiers are in Appendix~\ref{app:repro}.}; every number regenerates from committed artifacts.

\section{The Original Result and the Claims Under Test}
\label{sec:original}

\citet{lindsey2025biology} prompt Claude 3.5 Haiku with a couplet opening (``He saw a carrot and had to grab it,'') and study how the model completes the second line with ``rabbit''. Attribution graphs computed from a 30M-feature CLT localize the decision on the newline token before the second line: features for the candidate words ``rabbit'' and ``habit'' are active there, and, per their caption, ``the features are only active over the planning token''. Their Figure~13 then sweeps the position of a suppress-and-inject intervention (suppress the ``rabbit'' and ``habit'' features, inject a ``green'' feature found in a separate poem) across the prompt and plots the probability of predicting a different final token: ``the steering intervention only has an effect over the newline planning token''. In a separate test on a random sample of 25 poems, injecting a planned-word feature made the model end its line with the injected word in 70\% of cases. Two of their own caveats matter for what follows: no numerical steering strength is documented, and planned-word features were found ``in about half of the poems'', which the authors attribute either to incomplete CLT coverage or to the model not always planning.

\paragraph{Three claims.}
We decompose the result into three separately testable claims. \textbf{C1 (position specificity):} the intervention effect is concentrated at a single position; other positions are inert. \textbf{C2 (site identity):} that position is the newline preceding the line, upstream of the emitted word, rather than the position adjacent to emission. \textbf{C3 (plan representation):} features representing the planned word are active at the planning site before the line is written. C1 without C2 and C3 is compatible with improvisation: an intervention adjacent to emission that directly biases the output distribution. C2 and C3 are what make the result evidence of \emph{planning}. The distinction matters: \citet{jacopin2026prolepsis} describes its position sweeps as replicating Figure~13, but its effective site is the final prompt token, so it confirms C1 while, as we show below, C2 and C3 are not recovered.

\begin{table*}[t]
\centering\small
\begin{tabular}{@{}R{2.3cm}R{5.4cm}R{6.5cm}@{}}
\toprule
 & \citet{lindsey2025biology} & This work \\
\midrule
Model & Claude 3.5 Haiku (proprietary) & Gemma 2 2B, Llama 3.2 1B, Qwen3-0.6B, Qwen3-1.7B (open base models) \\
Transcoder & 30M-feature CLT (proprietary) & Six open CLTs: mntss ReLU (426K, 524K, 2.5M), BlueLightAI JumpReLU (16K, 20K per layer) \\
Feature discovery & Top-down: attribution graph from a specific completion; inject feature is a planned-word feature from another poem & Bottom-up: decoder-vector cosine against the vocabulary, CMU rhyme grouping; inject feature is the best decoder match for the alternative group or word \\
Intervention & Suppress planned-word features, inject alternative; one position at a time & Same design; strength swept over eight values (the original documents none) \\
Prompt geometry & The model composes the second line; steering upstream of the emitted word & Primary: prompt ends just before the rhyme word (next-token readout). Composition horizon (\S\ref{sec:horizon}): prompt ends at the newline, the model composes the line \\
Readout & Probability of predicting a different final token; on 25 poems, fraction of lines ending in the injected word (70\%) & Next-token probability of the injected word (absolute and ratio); fraction of 20 sampled lines ending in the inject group; greedy line \\
Breadth & One poem for Figure~13; 25 poems for the injection test & One prompt per cell for the grids; four prompts per reference cell; 444 pairs at fixed strength \\
\bottomrule
\end{tabular}
\caption{Protocol differences between the original steering-location experiment and this stress test. The feature-discovery row is the one that limits what our probes can show (\S\ref{sec:discussion}).}
\label{tab:protocol}
\end{table*}

\paragraph{What our probes can and cannot show.}
Table~\ref{tab:protocol} lists the protocol differences. Two are fixed throughout. First, bottom-up feature discovery biases discovered features toward output-adjacent (``say $X$'') behavior. A steering test built on such features can show that \emph{these} write-directions do not redirect from the newline; it cannot show that no newline-resident plan exists, because a plan represented in features that our scan never surfaces would be invisible to it. The census of \S\ref{sec:site} is designed to reduce this gap (it classifies every active feature, not only scan hits), but its plan-like criterion still reads decoders. Second, our primary sweeps use prompts that end immediately before the rhyme word; \S\ref{sec:horizon} runs the composition-horizon variant that matches the original's geometry. Throughout, a claim we ``do not recover'' is one that none of our probes finds; we reserve ``refute'' for nothing in this paper.

\section{Experimental Setup}
\label{sec:setup}

\paragraph{Models and transcoders.}
We use four open base models \citep{gemma2024,llama32024,qwen32025}, spanning two sizes of the same family (Qwen3-0.6B and 1.7B) and two architecture families (Llama 3.2 1B, Gemma 2 2B), crossed with six open CLTs from two independent pipelines: the \texttt{mntss} plain-ReLU CLTs (426K to 2.5M features) and the BlueLightAI JumpReLU CLTs (16K and 20K per layer) \citep{mntssClt2025,bluelightaiClt2026}. Crossing models, CLTs, and rhyme families yields the seven cells of Table~\ref{tab:cells}. CLTs are the cross-layer descendant of sparse-autoencoder dictionary learning \citep{templeton2024scaling}. The CLT implementation is validated against the Python Circuit Tracer reference \citep{circuittracer2025}: 90/90 top-10 features match with maximum relative error $1.2\times10^{-6}$.

\paragraph{Feature discovery.}
For each CLT we scan every feature's decoder vector against the full token-embedding matrix, keep features whose top tokens are clean English words, transcribe them with the CMU Pronouncing Dictionary \citep{cmudict}, and group them by rhyme ending (last stressed vowel onward). This yields 287 phonologically clean features for Gemma 426K (35 rhyme groups, 98 words), 79 for Llama 524K, and 636 to 739 for the Qwen3 CLTs.

\paragraph{Intervention protocol.}
Following the original's suppress-and-inject design: all discovered features of the prompt's natural rhyme group are clamped at strength $-s$, one feature of an alternative group is added at $+s$, both applied from the feature's source layer through all downstream layers, at a single token position. We sweep the position over every prompt token and the strength over $s\in\{0.5,1,2.5,5,10,25,50,100\}$ (a 2D grid), and measure the next-token probability of the injected word. Prompts are four-line completion poems ending in a trailing space (Appendix~\ref{app:prompts}); base models require this priming to rhyme at all (a 0\% rhyme rate on bare prompts rises to 78\% with priming).

\paragraph{Effect-size tiers.}
Because ratios over tiny baselines make weak effects look strong, every steering result is reported with its absolute probability and assigned to one of three tiers: \emph{behavioral} ($P\geq0.1$, the injected word competes for the greedy output), \emph{marginal} ($10^{-3}\leq P<0.1$), and \emph{logit-only} ($P<10^{-3}$, detectable in the distribution but with no effect on sampled text at these temperatures).

\begin{table}[t]
\centering\small
\begin{tabular}{@{}R{2.1cm}R{4.3cm}@{}}
\toprule
Experiment & Sample size \\
\midrule
Position-by-strength grids & 7 cells $\times$ 1 prompt $\times$ 20--32 positions $\times$ 8 strengths \\
Prompt breadth & 2 reference cells $\times$ 4 prompts, $s{=}25$ \\
Pair-level re-analysis & 3 CLTs $\times$ 4 prompts $\times$ 12--67 alternative groups = 136, 44, 264 pairs, $s{=}10$ \\
Newline census & 7 cells $\times$ 1 prompt $\times$ 6 positions, every active feature \\
Composition horizon & 4 cells, 1--4 prompts, 3 seeds, 4 conditions, 60 lines each: 36 runs, 8{,}640 lines, 36 sweeps \\
Newline patching & 7 pairs, 3 seeds, 60 lines each: 21 runs, plus 14 controls \\
Random controls & 3 cells $\times$ (10 random features + 10 random directions); 1 cell $\times$ 6 random-weight seeds \\
\bottomrule
\end{tabular}
\caption{Sample sizes per experiment. Single-prompt experiments are marked as such in the text; Limitations discusses why the prompt universe is small at this scale.}
\label{tab:n}
\end{table}

\section{Results}

\subsection{C1: one effective position, behavioral in two cells of seven}
\label{sec:localization}

\begin{figure*}[t]
\centering
\includegraphics[width=1\textwidth]{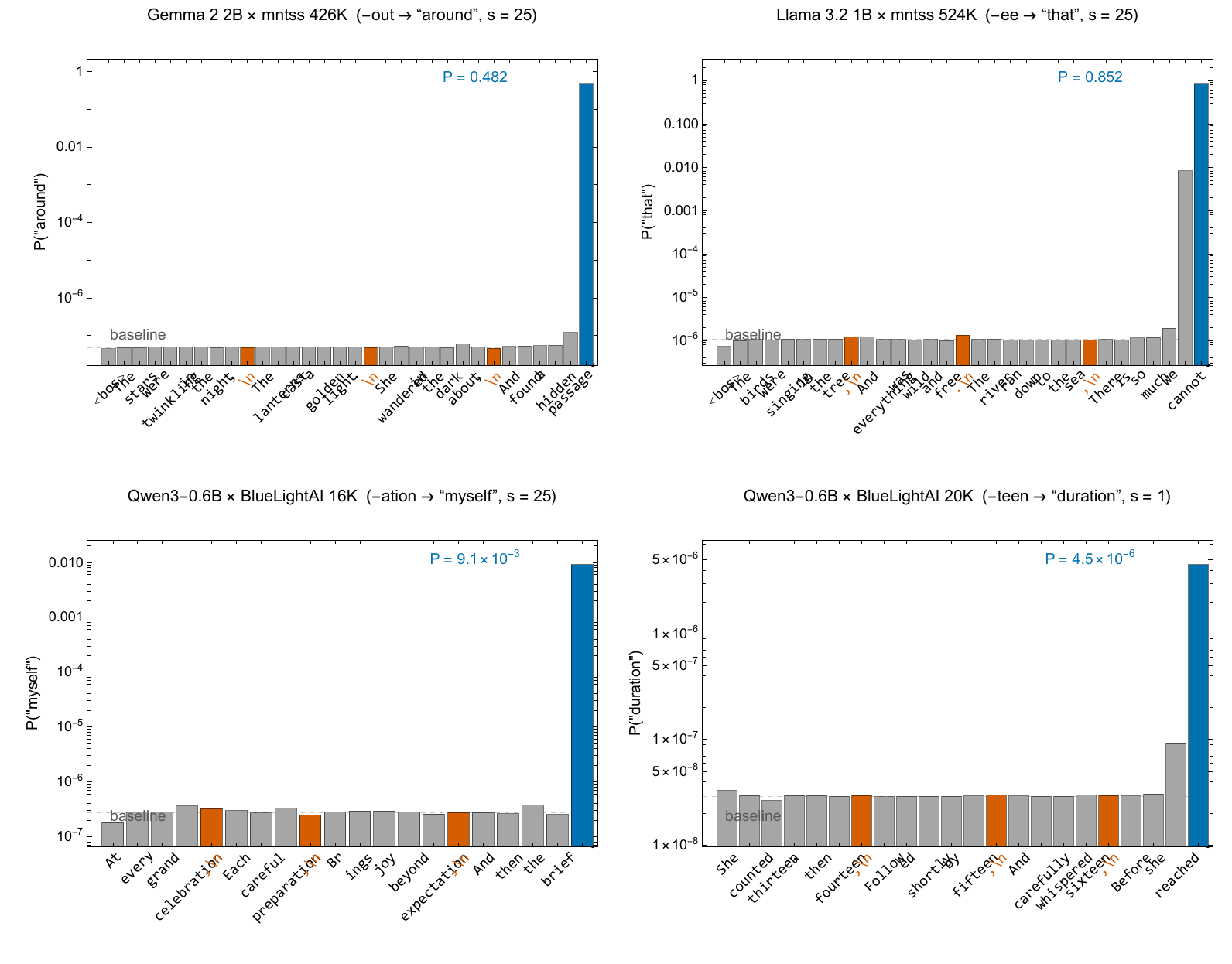}
\caption{Position sweeps for four cells (log scale), at each cell's best strength. Orange bars are newline tokens; the blue bar is the spike. The dashed line is the unsteered baseline probability of the injected word. The intervention is inert at every newline and effective only at the final prompt token, adjacent to emission. Note the vertical scales: the two top panels reach behavioral probabilities, the two bottom panels do not.}
\label{fig:sweep}
\end{figure*}

Table~\ref{tab:cells} summarizes the seven cells; Figure~\ref{fig:sweep} shows four position sweeps. In every cell the sweep is flat at baseline for every position except one, where it rises by a factor of $3.8$ to $10^{7}$ depending on the cell. Six of seven cells spike at the final prompt token; the seventh spikes one position earlier. The reference cells replicate and slightly exceed \citet{jacopin2026prolepsis}: Gemma $0.482$ at $s{=}25$ against $0.457$ at that work's fixed $s{=}10$; Llama $0.853$ against $0.777$.

The tiers of Table~\ref{tab:cells} qualify the headline. Only the two mntss cells are behavioral: the injected word reaches $0.48$ and $0.85$, enough to change generated text. The Qwen3-0.6B cell with the 16K development CLT is marginal ($0.009$), and the four Qwen3 cells with the production 20K CLTs are logit-only, moving the target from about $10^{-7}$ to at most $4.5\times10^{-6}$. The single-position \emph{signature} is present in all seven; \emph{redirection} in the sense of the original's 70\% is present in two.

\paragraph{Breadth.}
Two checks address the single prompt per cell. With each reference cell's intervention held fixed, the three other validated prompts of the earlier study, including prompts whose natural rime is foreign to the injected feature, all spike at the final token (8/8 prompts; Clopper-Pearson 95\% CI $[0.40,1]$ per cell; joint uniform-position null $7.2\times10^{-13}$). And we re-analyzed the committed position sweeps of \citet{pliprs2026}, run at a fixed $s{=}10$ over four prompts and every alternative rhyme group discovered for each CLT: 136 pairs on Gemma 426K, 44 on Llama 524K, 264 on Gemma 2.5M (Table~\ref{tab:pairs}). Of the 247 pairs whose best position exceeds ten times the sweep's own floor, 247 spike at the final token; all 28 pairs with a behavioral-tier effect do. The pairs whose maximum falls on a newline (29 of 444) are all flat profiles whose maximum is within $2.6\times$ of the floor, that is, noise. Under a uniform-position null for the seven grid cells (each spike equally likely at any of its $n_i\in[20,32]$ positions), all seven localizing within their last two positions has probability $\prod_i(2/n_i)=3.3\times10^{-8}$ (Appendix~\ref{app:newline}).

\begin{table*}[t]
\centering\footnotesize\setlength{\tabcolsep}{4.2pt}
\begin{tabular}{@{}llrllrrrll@{}}
\toprule
Model & CLT & Width & Act. & Rhyme & Baseline $P$ & Best $s$ & Best ratio & Best $P$ & Tier \\
\midrule
Gemma 2 2B & mntss & 426K & ReLU & -out & $4.8\times10^{-8}$ & 25 & $9{,}974{,}880\times$ & 0.482 & behavioral \\
Llama 3.2 1B & mntss & 524K & ReLU & -ee & $1.1\times10^{-6}$ & 25 & $806{,}260\times$ & 0.853 & behavioral \\
Qwen3-0.6B & BLA dev & 16K/layer & JumpReLU & -ation & $2.7\times10^{-7}$ & 25 & $33{,}860\times$ & $9.1\times10^{-3}$ & marginal \\
Qwen3-0.6B & BLA & 20K/layer & JumpReLU & -teen & $2.9\times10^{-8}$ & 1 & $157\times$ & $4.5\times10^{-6}$ & logit-only \\
Qwen3-1.7B & BLA & 20K/layer & JumpReLU & -teen & $1.1\times10^{-7}$ & 5 & $16.4\times$ & $1.8\times10^{-6}$ & logit-only \\
Qwen3-0.6B & BLA & 20K/layer & JumpReLU & -ation$^{\dagger}$ & $2.7\times10^{-7}$ & 10 & $5.4\times$ & $1.5\times10^{-6}$ & logit-only \\
Qwen3-1.7B & BLA & 20K/layer & JumpReLU & -ation & $2.1\times10^{-7}$ & 2.5 & $3.8\times$ & $8.2\times10^{-7}$ & logit-only \\
\bottomrule
\end{tabular}
\caption{The seven cells: models \citep{gemma2024,llama32024,qwen32025} $\times$ CLTs \citep{mntssClt2025,bluelightaiClt2026} $\times$ rhyme families. Best (strength, position) per cell from the full 2D grid; ratio is best $P$ over baseline $P$ of the injected word; tiers as defined in \S\ref{sec:setup}. BLA is BlueLightAI. Every cell spikes at the final prompt token except $^{\dagger}$ (one position before). In all seven cells the suppress and inject features are encoder-silent at the final prompt token (\S\ref{sec:site}), so at the effective site every cell tests bidirectional write-direction steering rather than suppression of an active feature. The word-level mntss 2.5M CLT (ReLU), used in \S\ref{sec:horizon} and Table~\ref{tab:pairs}, completes the open-CLT inventory.}
\label{tab:cells}
\end{table*}

\subsection{Is the final-token spike generic?}
\label{sec:generic}

A reviewer of an earlier version of this paper raised the right objection: the injected features are decoder-aligned, so injecting them at the final position gives a direct path to the next-token logits, whereas an earlier intervention must propagate through attention. The sweep may therefore identify the position of maximal causal leverage rather than the position at which the model commits to the rhyme. We test this directly with the per-position random controls (Figure~\ref{fig:random}, Appendix~\ref{app:newline}).

On the three cells above the logit-only tier, we replaced the inject feature with 10 layer-matched random CLT features or 10 norm-matched Gaussian directions (suppress side and strength unchanged) and recorded, at every position, the target's probability and, for random features, the probability of the drawn feature's own top decoder token. The location \emph{is} generic: random features spike their own token at the final position in 10/10 (Gemma), 9/10 (Llama), and 6/10 (Qwen) draws, by ratios of up to $10^{6}$ over their own tiny baselines; the remaining draws spike nowhere. The magnitude is not: the largest own-token probability any random draw reaches is $3.6\times10^{-6}$ (Gemma), $2.9\times10^{-3}$ (Llama), $2.3\times10^{-3}$ (Qwen), against $0.48$, $0.85$, and $0.009$ for the real feature. On the two behavioral cells the real feature exceeds the best random write-direction by two to five orders of magnitude; on the marginal Qwen cell it exceeds it by $4\times$ only, which is why we do not count that cell as evidence of feature-specific redirection. No random draw moves the \emph{target} above $7\times10^{-5}$ at any position.

Two further controls bound the triviality. For a feature whose source layer is the final layer, a final-token spike is architecturally forced (no attention follows; the logit readout sees only the last position). But the effect persists for features injected at middle layers: the strongest Gemma -out feature (layer 16 of 26, nine downstream attention layers) still concentrates its effect at the final token ($160{,}379\times$ over baseline under inject-only steering at $s{=}10$, though to an absolute probability of only $3.6\times10^{-4}$). And on a Gemma built from config with seeded random weights, or with every trained tensor's elements shuffled in place, the identical pipeline and features produce no spike (worst ratios $3.5\times$ and $99\times$ against $9{,}974{,}880\times$ trained; \S\ref{sec:ablations}).

We therefore restate C1 more carefully than the original's framing invites. The single-position signature says where a write-direction has the most leverage, and at this scale that is the position adjacent to emission. What is informative about planning is not the spike but its complement: the newline, which in Claude 3.5 Haiku was the \emph{only} effective position, is here inert (\S\ref{sec:site}), including when the model has a whole line to compose (\S\ref{sec:horizon}).

\subsection{C2 and C3: the newline is not recovered as a site}
\label{sec:site}

Every prompt contains three newline tokens, including the one immediately preceding the incomplete final line, the exact analogue of the original's planning site. Across all seven cells and all eight strengths, no newline position moves the target probability off baseline (Gemma 426K: $P\approx4.8\times10^{-8}$ at positions 9, 17, and 25, against $0.457$ at the final token under the identical $s{=}10$ intervention). The pair-level re-analysis extends this to 444 pairs: the best newline position never exceeds $1.4\times$ (Gemma 426K), $3.3\times$ (Llama), or $1.3\times$ (Gemma 2.5M) the sweep's floor (Table~\ref{tab:pairs}). The failure is not for want of causal reach: interventions at the newline have 6 to 25 subsequent token positions and, for mid-layer features, up to 9 downstream attention layers through which to propagate.

Steering with the \emph{wrong} features could still explain the inert newline: as \S\ref{sec:original} fixes, our discovery yields ``say $X$'' features by construction, while the original's attribution surfaced planned-word features resident on the newline. We therefore ran a census: at each newline, at mid-line control tokens, and at the final token, encode the residual through the CLT and classify \emph{every} active feature (not a top-$K$, so weak features cannot be truncated away) by whether its decoder points at the upcoming rhyme (two or more rhyme-group words in its top-20 vocabulary projection, or cosine $\geq0.3$ to the target word). The result is uniform across all seven cells: the plan-like rate at the newline never exceeds the mid-line base rate (Gemma: 0.00\% at newlines against 0.28\% at controls; Llama: 1.02\% against 2.06\%; the dense JumpReLU codes sit flat at 0.03\% to 0.76\% everywhere). Rhyme-decoding features fire at a constant background rate; nothing concentrates at the newline. So C3 is not recovered either: there is no CLT-visible representation of the upcoming rhyme at the newline to steer (full tables in Appendix~\ref{app:newline}). The census is one prompt per cell, and its criterion reads decoders, so a plan carried by features whose decoders do not surface rhyme words would evade it; see \S\ref{sec:discussion}.

\paragraph{At the effective site, every cell is write-direction steering.}
The census also records the encoder activation of the protocol's own steering features at each censused position (Appendix~\ref{app:newline}). In all seven cells, the suppress features and the inject feature are encoder-silent at the final prompt token, the position where the intervention works. Clamping a silent feature at $-s$ suppresses nothing; it injects a negative decoder direction. At the effective site the suppress-and-inject intervention is therefore bidirectional write-direction steering, decoupled from the model's actual feature activations, which explains mechanically why that site is emission-adjacent, and which means none of our cells tests the original's suppression of an \emph{active} planned-word feature. It also settles what the suppress half contributes in the inject-only ablation (\S\ref{sec:ablations}): a second write-direction, not the removal of an active feature. On the mntss cells the steering features are silent at every censused position. On the JumpReLU Qwen3 cells they are weakly active (0.05 to 0.29) at some mid-line control tokens, and on Qwen3-1.7B -ation two suppress features are active at newlines (15:263 at 0.11 and 0.53 on the second and third newlines, 18:4404 at 0.26 on the first), which is the one place in our data where a feature for the natural rhyme group is present at the original's planning site. The census counts it (it does not raise that cell's newline plan-like rate above control, Table~\ref{tab:census}), and the grid shows that clamping it there at any of the eight strengths leaves the target at baseline.

The census also caught our own reference code manufacturing the opposite conclusion: an earlier detection analysis in the implementation underlying \citet{jacopin2026prolepsis} read the encoder from the post-MLP residual and reported rhyme features active at the final token. Under the hook the CLT was actually trained on, those activations are exactly 0.000 (Appendix~\ref{app:lessons}).

\subsection{The composition-horizon test}
\label{sec:horizon}

\begin{figure*}[t]
\centering
\includegraphics[width=1\textwidth]{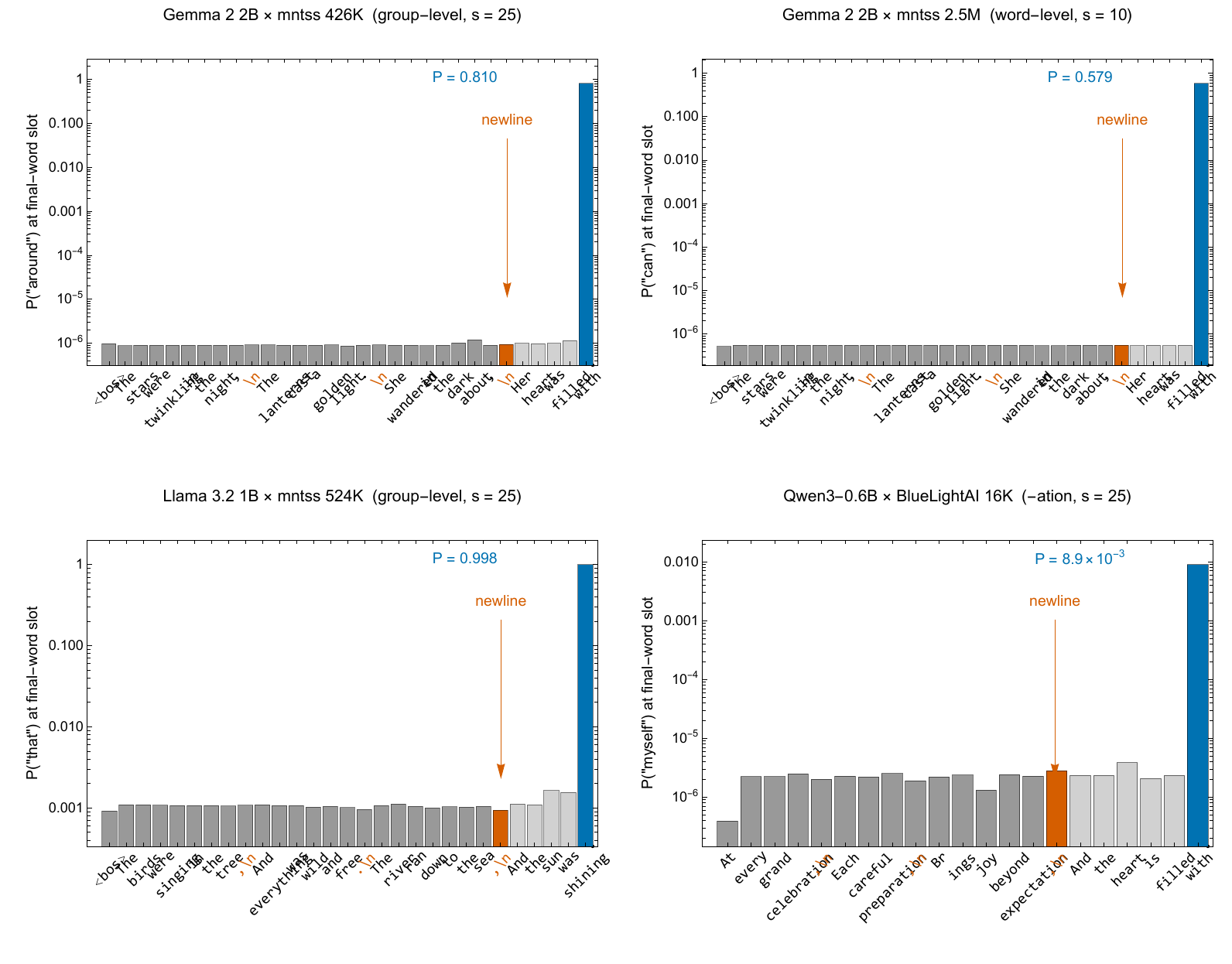}
\caption{Composition-horizon sweeps (log scale), one representative run per cell. The prompt ends at the line-3 newline (orange, arrow); lighter bars are the composed baseline line; $P(\text{inject word})$ is measured at the final-word slot. The newline sits at the sweep's floor for the rhyme target, including at word-level CLT resolution (top right); only steering inside the composed line, adjacent to emission, moves it. All 36 runs of the experiment agree: the peak falls inside the composed line in 36 of 36 and at the newline in 0 of 36, and the newline stays between 0.86 and 1.24 times its own floor.}
\label{fig:horizon}
\end{figure*}

The strongest remaining objection to the sweeps above is that prompts ending immediately before the rhyme word leave the model nothing to plan. We therefore rerun the experiment in the original's geometry: truncate the prompt at the newline ending line 3, so the model must compose the entire fourth line, and steer at that newline, which is now also the final prompt token, handing the newline-time signal its best possible chance. Per condition (baseline, suppress-only, inject-only, suppress-and-inject) we record sampled lines (temperature 0.7) classified by the CMU rime of their final word, the greedy line, and the teacher-forced probability of the inject word at the final-word slot, swept over every steering position (Figure~\ref{fig:horizon}).

Four cells carry this test: the two behavioral cells, the marginal Qwen3 cell, and the word-level 2.5M-feature Gemma CLT, whose features are the closest open analogue to the ``ordinary features representing the planned word'' of the original. Because an earlier version of this experiment used one prompt and 20 lines per condition, which cannot separate a near-zero baseline from anything below 40\%, we reran it at the scale its conclusion requires: every validated prompt per cell (four on the Gemma cells, three on Llama, one on Qwen3) crossed with three sampling seeds, 60 lines per condition per run, 36 runs and 8{,}640 sampled lines in total (Table~\ref{tab:n}). Decision criteria were registered before the runs and are reproduced in Appendix~\ref{app:newline}; all analysis scripts and per-line outputs are in the repository.

\paragraph{The newline moves the next token, and only the next token.}
The intervention at the newline is not weak; it is short-ranged. It captures the token immediately after the steering site: the injected word becomes the \emph{first word} of the composed line in 703 of 720 sampled lines on Gemma 426K under suppress-and-inject, against 0 of 720 at baseline, and in 470 of 540 on Llama against 1 of 540 (Table~\ref{tab:horizon}). Of the 1{,}545 lines across all four cells in which the injected word appears under suppress-and-inject, it is the line's first word in 1{,}540. What it does not do is survive to the end of the line. Six words later, at the rhyme, the same intervention has left no trace.

This is the emission-adjacent mechanism of \S\ref{sec:generic} seen end on. In this geometry the newline \emph{is} the final prompt token, so steering there biases the next-token distribution exactly as it does in the primary sweeps; the composed line then proceeds from that word under the model's own dynamics. The position sweeps make the same point from the other side. Across all 36 runs the sweep peaks at a position inside the composed line, never at the newline (36/36 and 0/36), and the newline sits at its own floor in every run, at a ratio of 0.86 to 1.24 (median 0.99 on Gemma 2.5M, 1.02 on Gemma 426K, 0.88 on Llama). The peak reaches a median $P(\text{inject})$ of 0.88 on Gemma 426K, 0.74 at word level and 0.999 on Llama, against a newline median of $6.7\times10^{-5}$, $1.1\times10^{-6}$ and $6.4\times10^{-4}$.

\paragraph{The rhyme is never redirected.}
Under the registered metric, the fraction of composed lines whose final word falls in the injected rhyme group, no cell separates from its baseline: Gemma 426K 38/720 against 21/720, Llama 4/540 against 0/540, Qwen3 2/180 against 0/180, Gemma 2.5M 0/720 against 0/720, with overlapping intervals everywhere (Table~\ref{tab:horizon}). H1 fails in all four cells.

That metric is confounded, and in the direction that flatters the planning hypothesis: a line that begins with ``around'' and ends with ``around'' or ``found'' is counted as redirected. We therefore recount the inject-group endings under criteria that exclude lines into which the word was inserted, from the registered metric through to ``the injected word does not occur in the line at all'' (Table~\ref{tab:sensitivity}). The nominal rise on Gemma 426K, the only cell with any counts to speak of, survives no control: 21 to 38 raw ($p=0.03$), 17 to 23 once lines ending on the injected word itself are removed ($p=0.42$), and then it \emph{reverses}, to 20 against 4 ($p=1.4\times10^{-3}$) and to 20 against 0 under the two strictest criteria ($p=1.7\times10^{-6}$ and $1.4\times10^{-5}$). Steering does not add genuine redirects; it removes the ones the model produces on its own. At baseline Gemma 426K ends 20 of 720 lines cleanly in the inject rhyme group, with well-formed completions such as ``Her father was nowhere to be found.''; under either condition that injects, it ends none. On the other three cells the count of genuine redirects under steering never exceeds two under any criterion, and under the strictest one it is zero in every cell but for a single Qwen3 line.

This recount was not pre-registered and we report it as exploratory; Table~\ref{tab:sensitivity} exists so the reader can apply whichever criterion they consider fair, including the registered one. The conclusion is the same under all six.

\paragraph{What the design can exclude.}
Overlapping intervals do not show equivalence, so we state the resolution. Pooled over prompts and seeds, each condition holds 720 lines on the Gemma cells, 540 on Llama and 180 on Qwen3. The smallest redirect fraction whose interval would clear the baseline's is 0.013 on Gemma 2.5M, 0.017 on Llama, 0.050 on Qwen3 and 0.061 on Gemma 426K, the last because its baseline is a non-zero 21/720: the prompts' natural group (AW1~T, ``about'') and the injected group (AW1~N~D, ``around'') share a nucleus, so the model lands in the injected group spontaneously. Every cell therefore excludes an effect of the size the original reports, 70\% of lines on 25 poems, by a very wide margin, and the three cells with near-zero baselines exclude anything above about 2\%. What we cannot exclude is a redirect rate below one or two percent.

\paragraph{Composition is derailed, not degraded.}
Two further checks separate the mechanism from two plausible alternatives. Newline injection does not simply break the model: the share of lines repeating a word three or more times is statistically unchanged on Gemma 426K (2.6\% to 3.5\%, $p=0.36$) and Llama (0.4\% to 0.2\%), and \emph{falls} on Gemma 2.5M (2.6\% to 0.7\%, $p=6\times10^{-3}$), while the type-token ratio moves by at most 0.02 and upward, toward more varied lines, in three cells of four. Lines do shorten on Gemma 426K, from 7.4 to 6.1 words. Nor is the natural rhyme destroyed: the registered contrasts on the natural-group rate give one result below the corrected threshold of $\alpha=0.0042$, inject-only on Gemma 426K (220/720 to 164/720, $p=1.0\times10^{-3}$), whose sign is inconsistent across the four prompts; the drop reported in the earlier version of this paper on Llama does not replicate at scale ($p=0.048$). H2 therefore fails as a claim about plan disruption: what the numbers describe is a line that begins elsewhere, not a plan that has been disturbed. The typical steered line is ``Around the beach he soon found out.'', which starts on the injected word and still ends in the prompt's own rhyme group.

Across three model families, two CLT pipelines, both feature granularities, 36 runs and 8{,}640 composed lines, the verdict is uniform: steering at the newline rewrites the beginning of the line and leaves its rhyme untouched, while the only position that redirects the rhyme is the one adjacent to emission. This is consistent with \citet{hanna2026latent}, whose smallest models show the weakest planning signal, and with the collapse of rhyme rates under forced non-default targets that we report in Limitations.

\subsection{Patching the newline, without a transcoder}
\label{sec:patching}

Every probe so far writes a CLT decoder direction, so every null can be blamed on
feature discovery (\S\ref{sec:original}). Activation patching cannot be: it
\emph{replaces} a residual row with one the model itself produced, needing no
features, no transcoder and no strength. We therefore ran it as the decisive test
of reading (ii).

\paragraph{Design.}
Donor and recipient are a \emph{minimal pair}: two prompts identical through line
3 except that line's final word, which sets a different rhyme. Patching the last
prompt position transfers its whole state, so with arbitrary prompts a rhyme
change would only show that the recipient inherited the donor's context; with a
minimal pair, everything the patch can carry is shared except the rhyme. The
donor's row is patched into the recipient's line-3 newline, one layer at a time
and at every layer at once, and the model then composes line 4. Seven
prompt-direction pairs on Gemma 2 2B $\times$ three seeds give 1{,}260 sampled
lines per condition. Criteria were registered before the runs (Appendix~\ref{app:newline}).

Only Gemma carries this test. The design needs AABB prompts, where line 3 opens a
new couplet, and Llama 3.2 1B rhymes on those in 0 to 8\% of lines against Gemma's
20 to 42\%, below the 15\% its own baseline would need for a redirect to be
detectable. That is consistent with its validated prompts being monorhyme and with
prior reports that Llama rhymes less reliably than Gemma \citep{pliprs2026}. The
result is therefore single-model, and we report it as such.

\paragraph{The newline does not carry the rhyme.}
Replacing the newline's entire state, at every layer, with that of a poem ending
on a different rhyme moves the composed line's rhyme in 11 of 1{,}260 lines against
4 at baseline (Fisher $p=0.12$); one layer at a time, 10 of 1{,}260 ($p=0.18$).
The design resolves a redirect rate of 1.4\%. The recipient's \emph{own} rhyme is
equally untouched, 384/1{,}260 against 357/1{,}260 ($p=0.26$). The patch is not
inert on the text: the composed line opens with the donor's line-3 word 33 times
against 4 at baseline ($p=9\times10^{-7}$), the same next-token capture as
\S\ref{sec:horizon}, at 2.6\% rather than 98\% because a minimal-pair patch
carries so little.

That "so little" is itself measured, and it is not nothing: the donor's and
recipient's newline rows differ down to cosine 0.925. The newline does encode
which word line 3 ended on. It is what the model does with that difference,
namely nothing to the rhyme six words later, that this experiment establishes.

\paragraph{Controls.}
Patching a prompt with its own row is a bit-exact no-op in all 42 runs (maximum
probability shift $1.4\times10^{-8}$). A donor whose line 3 ends in a CMU
rime-mate of the recipient's does not move the rhyme (136/420 to 114/420,
$p=0.11$), so the instrument does not move rhymes for reasons unrelated to the
plan. Patching at a mid-line position instead, where a minimal pair's rows are
identical by construction, is an exact no-op (136/420 to 136/420, $p=1.000$);
that null is evidence the harness writes what it claims, and an earlier version
that failed it exposed a real defect (Appendix~\ref{app:lessons}).

\subsection{Magnitude is a property of the transcoder}
\label{sec:magnitude}

Holding the protocol fixed, the best redirect ratio varies by more than six orders of magnitude across CLTs (Table~\ref{tab:cells}), and the ordering follows the transcoder, not the model. Both mntss cells sit at $10^{5}$ to $10^{7}\times$ with absolute probabilities of 0.48 to 0.85; the BlueLightAI production cells sit at $3.8$ to $157\times$ with absolute probabilities that never exceed $5\times10^{-6}$. The 16K development CLT lands in between ($33{,}860\times$) for a reason instructive in itself (\S\ref{sec:ablations}). A study that had tested only Qwen3 with the production 20K CLTs would have reported the steering result as marginal; one that had tested only Gemma with mntss would have called it dramatic. Neither would have been wrong about its own cell.

The evaluation metric compounds this. The $33{,}860\times$ cell has an absolute redirect probability of $0.009$; the $157\times$ cell, $4.5\times10^{-6}$. Ratios over tiny baselines make weak effects look strong, and the original reports neither a ratio nor a baseline, only an intervention-condition probability. We recommend reporting absolute probability, baseline, and a behavioral readout, as Table~\ref{tab:cells} and \S\ref{sec:horizon} do.

\subsection{No monotonic scaling within Qwen3}

Within the matched pair (same CLT pipeline, same width, same prompts, same protocol), the 0.6B model shows a \emph{stronger} final-token spike than its 1.7B sibling on both rhyme families: $157\times$ against $16.4\times$ (-teen) and $5.4\times$ against $3.8\times$ (-ation), all four in the logit-only tier. Feature availability also favors the smaller model (739 against 636 phonologically clean features). This does not contradict the 0.6B-to-14B trend of \citet{hanna2026latent}, which we cannot test above 1.7B on consumer hardware, but it shows the within-family trend is not monotonic at its lower end, and that CLT training quality is a plausible confound in any cross-size comparison.

\section{Ablations and Controls}
\label{sec:ablations}

\paragraph{Steering strength.}
The original documents no numerical steering strength; the $s{=}10$ used by \citet{jacopin2026prolepsis} is that work's internal convention. Figure~\ref{fig:random} (bottom right, appendix) shows why this free parameter deserves a grid: the optimum is $s{=}25$ on both mntss reference cells, $s{=}5$ on Qwen3-1.7B (-teen), and $s{=}1$ on Qwen3-0.6B (-teen), where the response declines monotonically with strength. Three qualitative regimes appear (rise-plateau, rise-decline, monotone decline), so a single global strength misstates every cell but one.

\paragraph{Inject-only versus suppress-and-inject.}
Because CLT decoder vectors are residual-stream directions, injecting one is additive steering; the question is what the suppress side adds. Rerunning three Qwen3 cells with suppression disabled: on Qwen3-0.6B -teen the matched inject-only run retains 96\% of the full effect ($47.5\times$ against $49.5\times$), on -ation one third ($1.83\times$ against $5.4\times$), and on Qwen3-1.7B -teen one fourteenth ($1.21\times$ against $16.4\times$). The decoder-as-steering-vector mechanism is real, and the suppress side ranges from nearly redundant to essential across cells; neither half of the protocol can be dropped in general. In every cell the suppress side is, by the census, a second write-direction at the steered position rather than the removal of an active feature (\S\ref{sec:site}).

\paragraph{Choice of inject feature.}
The single largest lever we found is which feature is injected. On Qwen3-0.6B with the 16K development CLT, replacing the rhyme group's top feature with the feature most decoder-similar to the specific target word moved the cell from $3.5\times$ to $33{,}860\times$, four orders of magnitude, and from that cell alone one would conclude the 16K CLT is a far better steering substrate than the wider 20K production CLT. On the -teen cells the \emph{opposite} pick wins: the group-broad feature beats the word-specific one by $3\times$. Published steering results that rest on one hand-picked feature per condition, as the original's do and as ours would absent this ablation, carry a variance that the papers do not report.

\paragraph{Random baselines.}
Three controls in the dead-salmon lineage \citep{meloux2025dead}, all pre-registered (Appendix~\ref{app:newline}). \textbf{Random features and directions} are analyzed per position in \S\ref{sec:generic}; in summary, the target's absolute probability stays below $7\times10^{-5}$ in all 60 draws, against 0.48 to 0.85 (mntss) and 0.009 (Qwen 16K) for the real feature. One amendment: our pre-registered criterion was a $10\times$ \emph{ratio} bound, and only Gemma passes it; on the common-token targets (`` that'', `` myself'') many random draws lift the target $10$ to $200\times$ over its tiny baseline, because generic emission-site steering lifts frequent tokens. This is the decoder-direction regime of \S\ref{sec:generic}, and it re-teaches \S\ref{sec:magnitude}'s lesson: over tiny baselines, report absolute probabilities. \textbf{Random model:} on a Gemma built from config with seeded random weights, and on a stricter variant with every trained tensor's elements shuffled in place (preserving all norm statistics), the identical pipeline and features produce no spike: worst ratios $3.5\times$ and $99\times$ against $9{,}974{,}880\times$ trained, seed-unstable, one shuffle seed peaking off the final token. The machinery cannot manufacture the effect on a network without learned structure; it can, as \S\ref{sec:generic} shows, manufacture the \emph{location} on a trained one.

\section{Alternative Lenses}
\label{sec:baselines}

\paragraph{Non-CLT baseline: contrastive steering.}
\citet{maar2026plan} steer rhyming behavior with raw mean-difference directions and report effects across 23 models. Replicating their exact protocol (their prompts, layer $\lfloor0.8n\rfloor$, multiplier $m{=}1.5$, last-token hook) reproduces their headline on Llama 3.2 3B: rhyme rate 60\% to 30\%, all six flips hit-to-miss. But the direction norm $\lVert d\rVert$ varies tenfold across architectures (4.0 on Llama 1B, 11.5 on Llama 3B, 116.1 on Gemma 2B), so a global $m$ lands at very different effective perturbations; and the effect \emph{direction} flips by family (Gemma 2 2B is enhanced, $+20$\,pp at $m{=}1.0$, rather than inhibited) even at matched effective perturbation $m\lVert d\rVert$. Contrastive steering thus shifts rhyming behavior at the family level without word-level control, and its cross-model comparisons inherit an unnormalized-scale confound.

\paragraph{Per-layer transcoders.}
Final-token steering is not CLT-specific: on Llama, a per-layer transcoder (PLT) under a method-matched top-5 protocol spikes at $\Delta P=+0.986$. The same protocol initially scored the CLT arm at noise ($\Delta P\approx5.7\times10^{-7}$) until the feature-ranking criterion was corrected (maximum over target layers, not same-layer), after which the CLT recovered $\Delta P=+0.87$ to $0.92$. On Gemma the method-matched protocol fails for both transcoder types even though hand-picked features redirect at $P=0.457$: a ranking criterion alone flips the qualitative conclusion of a transcoder comparison.

\section{Discussion}
\label{sec:discussion}

\paragraph{Verdicts.}
\textbf{C1 generalizes,} with a qualification: single-position causal specificity appears in 7/7 cells and 247/247 detectable pairs, but the location is the generic leverage point of any write-direction; only its magnitude is feature-specific, and only two cells reach behavioral magnitudes. \textbf{C2 and C3 are not recovered:} the newline carries no census-visible plan at group- or word-level resolution, and is inert \emph{for the rhyme target} across 7 cells, 444 pairs and 8{,}640 composed lines. It is not inert in general, which is the sharper form of the result: under a composition horizon it captures the next token almost deterministically and decays to nothing within the line, so the effective site is emission-adjacent even when the intervention is applied a whole line upstream. \textbf{Boundary conditions:} magnitude is CLT-bound across six orders of magnitude; optimal strength varies $25\times$ across cells; one feature choice can dominate a cell; within-family scaling is not monotonic at the low end.

\paragraph{Three readings, and which we can exclude.}
(i) \emph{Scale.} The original's model is two to three orders of magnitude larger than ours, and \citet{hanna2026latent} report planning strengthening with scale. On this reading our result is a boundary condition: below some size between 2.6B and Claude 3.5 Haiku, the rhyme is committed at emission. Our data are fully consistent with it and cannot test it. (ii) \emph{CLT-invisible planning.} A plan could be represented in a distributed or abstract form that neither our decoder scan nor our census criterion surfaces, and that our ``say $X$'' inject features cannot engage from the newline; the original itself found planned-word features in only about half of its poems. This reading is now tested directly (\S
ef{sec:patching}) and loses its main support: an instrument that uses no features, no transcoder and no steering strength finds nothing at the newline either. What our feature-based probes establish is narrower, and the composition-horizon runs sharpen it: every write-direction we could find, at every strength, at every newline position, in every cell and pair, leaves the rhyme target at its floor, while the same directions redirect it at emission. The horizon runs add that this is not a failure of the intervention to take hold. It takes hold with near-certainty on the token it abuts and has decayed to nothing six words later, which is what a plan, on the original's account, is precisely supposed to survive. (iii) \emph{Improvisation.} The tested models choose the rhyme at emission. This is the simplest account of (i) and (ii) together, and it is the account the original's own dichotomy assigns to an emission-adjacent effective site, but it is an inference from absence, and we no longer state it as a finding. The earlier version of this paper did; the reviewers were right that the evidence supports ``not recovered'' and not ``refuted''.

\subsection*{}
\paragraph{What we did not reproduce, and what would.}
It is worth stating plainly what this paper is not. We reproduced Figure~13's \emph{shape} on open models, and we can say with some precision where the intervention acts. We did not reproduce its \emph{mechanism}: no probe we have run recovers the newline-resident planned-word features the original describes, and on these models the internals behind the figure remain out of reach. Getting the curve is not the same as getting the circuit, and a reproducibility study should say which one it has.

The question we would most like answered is therefore still open: \emph{what is the smallest model in which the whole of the original's result replicates}, features at the newline included, rather than only its causal signature? That is a search over model scale, and it needs models above 2.6B with either an open transcoder or an instrument that does not need one. \S
ef{sec:patching} shows the second exists: activation patching runs on any model that fits the hardware. We regard locating that floor as the natural successor to this work.

\paragraph{Reproducibility lessons.}
(1) The original reports no steering strengths, no baseline probabilities, and its figures are interactive rather than numeric, so ``replication'' requires reverse-engineering free parameters. (2) Feature discovery method is load-bearing: bottom-up decoder-similarity scans find emission-adjacent features, attribution finds planning features, and the two lead to different scientific conclusions from the same intervention machinery. (3) Ratios over tiny baselines, and CI overlap read as equivalence, both mislead; report absolute probabilities, a behavioral readout, and the minimum effect the design can detect. (3b) A behavioral readout can be confounded by the intervention that produces it. Our rhyme-class metric counts a line as redirected when it \emph{ends} in the injected group, and steering puts the injected word at the line's \emph{start}, so lines that merely repeat it score as successes; the sign of the effect flips once that is controlled (Table~\ref{tab:sensitivity}). Any steering study whose success metric can be satisfied by the injected token itself should report the same recount. (4) Encoder hooks and tokenizers are silent failure modes (Appendix~\ref{app:lessons}).

\section*{Limitations}

Our cells span 0.6B to 2.6B parameters; the original's subject model is far larger, so scale alone may explain the site relocation, and our results say nothing about Claude 3.5 Haiku itself. Feature discovery is bottom-up, so our steering and census probes can only see plans that surface through decoder vectors; a distributed or abstract plan would evade them (\S\ref{sec:discussion}). The census uses one prompt per cell. The composition-horizon test uses every validated prompt per cell, but that is four prompts on the Gemma cells, three on Llama and one on Qwen3, so its breadth is bounded by the competence limit described below rather than by sampling; it excludes redirect rates above 1.3 to 6.1\% and says nothing below that. Its redirect recount (Table~\ref{tab:sensitivity}) is post hoc, and although the conclusion holds under all six criteria including the registered one, a criterion chosen after seeing the data cannot carry the weight of a pre-registered one. The next-token sweeps carry per-prompt breadth on the reference cells (4/4 each) and pair-level breadth across 444 pairs, but the latter were run at a fixed $s{=}10$ by earlier code \citep{pliprs2026} whose recorded baseline field we found unreliable, so we measured newline effects against each sweep's own floor (Appendix~\ref{app:newline}). Wider prompt sets are bounded by model competence rather than by budget: at this scale, base models rhyme only within a narrow priming regime (78\% on validated completion prompts at best), forcing non-default rhyme words collapses rhyme rates to 1 to 26\%, and non-poetry planning domains we attempted (gridworld navigation) fail at task competence before planning is testable. We report this as a boundary condition of the phenomenon, not a sampling choice. In every cell the suppress features are encoder-silent at the effective site, so our cells test write-direction steering rather than the original's suppression of an active planned-word feature. The composition-horizon harness recomputes the forward pass per generated token and re-applies the newline hook each step; a KV-cache implementation should be checked to agree. The activation-patching test of \S
ef{sec:patching} runs on Gemma 2 2B alone, because Llama does not rhyme reliably on the AABB prompts its minimal-pair design requires; it is a single-model result and does not speak to the other three cells. The Maar replication covers 3 of their 23 models.

\section*{Acknowledgments}
An earlier version of this paper was submitted to the BlackboxNLP 2026 special track on reproducibility and reliability in interpretability analyses. The three anonymous reviewers' comments, in particular on effect-size reporting, on the structural bias of the position sweep, and on the power of the composition-horizon test, shaped this revision; the new analyses of \S\ref{sec:generic}, the pair-level re-analysis, and the rerun composition-horizon experiment of \S\ref{sec:horizon} are direct responses. The insertion mechanism reported there was found only because a reviewer asked for the power calculation that made the rerun necessary. Claude Code (Anthropic) assisted with prose drafting, LaTeX, and the analysis scripts over committed artifacts; all experiments, scientific content, and conclusions are the author's, who takes full responsibility.

\bibliography{references}

\appendix

\section{Prompts and Feature Identifiers}
\label{app:prompts}

Four-line completion prompts, one per cell family; the final line is incomplete and the prompt ends with a trailing space. Example (Gemma, natural group -out, inject ``around''):

\begin{quote}\small\ttfamily
The stars were twinkling in the night,\\
The lanterns cast a golden light.\\
She wandered in the dark about,\\
And found a hidden passage \textvisiblespace
\end{quote}

Table~\ref{tab:features} lists the suppress and inject feature identifiers (layer:index) for every cell. Suppression uses the natural rhyme group's top features by decoder cosine; on the Qwen3 -ation cells the inject feature is the one most decoder-similar to the specific target `` myself'', while the -teen cells inject a cluster-broad -ation feature (the ablation of \S\ref{sec:ablations} quantifies this choice).

\begin{table}[h]
\centering\small
\begin{tabular}{@{}lR{2.9cm}l@{}}
\toprule
Cell & Suppress & Inject \\
\midrule
Gemma 426K & 16:13725, 25:9385 & 22:10243 \\
Gemma 2.5M & 25:57092, 23:49923, 20:77102 & 25:82839 \\
Llama 524K & 13:30985, 9:5488, 14:27874, 13:32049 & 14:13043 \\
Q3-0.6B 16K -ation & 23:11154, 20:10987, 14:10719 & 22:8011 \\
Q3-0.6B 20K -ation & 19:9578, 0:8867, 25:4979 & 22:4081 \\
Q3-0.6B 20K -teen & 27:16425, 23:15839, 26:6308 & 19:9578 \\
Q3-1.7B 20K -ation & 15:263, 18:3801, 18:4404 & 21:3908 \\
Q3-1.7B 20K -teen & 27:16975, 20:3668, 18:10986 & 15:263 \\
\bottomrule
\end{tabular}
\caption{Feature identifiers (layer:index) per cell: the seven cells of Table~\ref{tab:cells} plus the word-level Gemma 2.5M cell of \S\ref{sec:horizon}. Q3 is Qwen3.}
\label{tab:features}
\end{table}

\section{Reproduction Pipeline}
\label{app:repro}

Each cell reproduces with three commands: pre-cache model and CLT; run the vocabulary scan and CMU filter; run the position-by-strength grid. Wall-clock on an RTX 5060 Ti: scan 3 to 4 minutes, grid seconds per strength, downloads 17 to 41\,GiB one-time. The repository \citep{paperrepo2026} is organized in three tiers: every committed artifact behind every number in this paper (tier 0), a Python analysis layer that re-derives each table and statistic from those artifacts with no GPU (tier 1, minutes), and the full Rust harness with a source snapshot of the underlying library \citep{candlemi2026}, which type-checks as staged, for end-to-end regeneration (tier 2). The pair-level sweeps of Table~\ref{tab:pairs} are the committed outputs of \citet{pliprs2026}. Gemma 2 and Llama 3.2 are gated models: regeneration requires accepting their licenses on HuggingFace.

\section{Registered Predictions and Full Results}
\label{app:newline}

The census and composition-horizon experiments were specified, with decision criteria registered, before any run. Registered outcomes: \emph{improvisation} (no rhyme-anticipating features at the newline; no newline condition shifts the composed line's final word) against \emph{wrong features} (plan-like features found at the newline; steering them redirects the line), with \emph{partial} reserved for a degraded natural-rhyme rate without inject-group gains. The observed outcome is the first in all cells for the rhyme target, plus a \emph{decoder-only regime} at the effective site in every cell (the steering features are encoder-silent at the final token, so suppression was anti-injection), and two uncorrected natural-rate drops that match the \emph{partial} pattern but do not survive multiple-comparison correction (\S\ref{sec:horizon}).

\paragraph{Census (Experiment 1).}
Plan-like rate (plan-like / active features), all active features classified, encoder hook validated by MLP-output reconstruction (Table~\ref{tab:census}).

\begin{table}[h]
\centering\small
\begin{tabular}{@{}lrrr@{}}
\toprule
Cell & newline & control & final \\
\midrule
Gemma 426K (-out) & 0.00\% & 0.28\% & 0.00\% \\
Llama 524K (-ee) & 1.02\% & 2.06\% & 2.78\% \\
Qwen3-0.6B 16K (-ation) & 0.38\% & 0.41\% & 0.32\% \\
Qwen3-0.6B 20K (-teen) & 0.026\% & 0.030\% & 0.066\% \\
Qwen3-1.7B 20K (-teen) & 0.050\% & 0.042\% & 0.064\% \\
Qwen3-0.6B 20K (-ation) & 0.545\% & 0.518\% & 0.514\% \\
Qwen3-1.7B 20K (-ation) & 0.758\% & 0.692\% & 0.690\% \\
\bottomrule
\end{tabular}
\caption{No cell shows newline enrichment; the rate is flat or lowest at the newline. Absolute counts differ by CLT density (hundreds of active features per position for ReLU mntss, tens of thousands for JumpReLU BlueLightAI), which the rate comparison controls for.}
\label{tab:census}
\end{table}

\paragraph{Steering-feature activity.}
Encoder activation of each cell's suppress and inject features (Table~\ref{tab:features}) at the six censused positions, read from the same census files. At the final prompt token: silent in all seven cells. At the three newlines: silent in six cells; on Qwen3-1.7B -ation, suppress feature 15:263 at 0.11 and 0.53 (second and third newlines) and 18:4404 at 0.26 (first newline). At the two mid-line controls: silent on the mntss cells; on the Qwen3 cells, one steering feature active at one control token in four of five cells (0.05 to 0.29), plus 21:3908 (inject) at 1.69 on Qwen3-1.7B -ation at the token ending ``Brings''. The reference cells' densities differ by two orders of magnitude (179 to 390 active features per position for mntss, 8{,}735 to 55{,}116 for BlueLightAI), so a chance hit is far likelier on the Qwen3 cells.

\paragraph{Composition-horizon steering (Experiment 2).}
Table~\ref{tab:horizon} gives, for every cell and condition pooled over prompts and seeds, the final-word rime classes of the composed lines, how often the injected word was inserted into the line and where, and the natural-group rate. Table~\ref{tab:sensitivity} gives the redirect recount of \S\ref{sec:horizon}. Registered contrasts are the twelve natural-rate tests (four cells by three conditions) at $\alpha = 0.05/12 = 0.0042$; the recount is exploratory. The bridge experiment (steering census-identified newline features) was not triggered: the census surfaced nothing to steer. Runs used seeds 1, 2 and 3 at 60 lines per condition; before them, the four cells were rerun at the earlier setting (one prompt, 20 lines, seed 42) on the rebuilt binary and reproduced the earlier results exactly, with identical greedy lines and identical class counts in all sixteen conditions, so no stack drift separates the two.

\begin{table*}[t]
\centering\footnotesize
\begin{tabular}{@{}llrrrrrp{4.6cm}@{}}
\toprule
 & & \multicolumn{3}{c}{final-word rime class} & \multicolumn{2}{c}{inserted} & \\
\cmidrule(lr){3-5}\cmidrule(lr){6-7}
Cell ($s$, $n$/cond.) & Condition & nat. & inj. & other & any & 1st & Greedy line \\
\midrule
\multirow{4}{*}{Gemma 426K (25, 720)} & baseline & 220 & 21 & 479 & 5 & 0 & Her heart was filled with doubt. \\
 & suppress-only & 217 & 17 & 486 & 5 & 0 & A smile on her face, a heart full of doubt. \\
 & inject-only & 164 & 31 & 525 & 639 & 638 & Around the town, she went about. \\
 & suppress+inject & 188 & 38 & 494 & 703 & 703 & Around the town, she went about. \\
\midrule
\multirow{4}{*}{Gemma 2.5M (10, 720)} & baseline & 220 & 0 & 500 & 9 & 0 & Her heart was filled with doubt. \\
 & suppress-only & 239 & 0 & 481 & 10 & 0 & The moon was shining bright. \\
 & inject-only & 219 & 0 & 501 & 358 & 355 & Can't find her way back to the house. \\
 & suppress+inject & 244 & 0 & 476 & 370 & 366 & Can't find her way back to the house. \\
\midrule
\multirow{4}{*}{Llama 524K (25, 540)} & baseline & 61 & 0 & 479 & 19 & 1 & And the sun was shining bright \\
 & suppress-only & 41 & 0 & 499 & 16 & 3 & Our hearts were filled with joy \\
 & inject-only & 71 & 1 & 468 & 479 & 478 & That was the way to go \\
 & suppress+inject & 70 & 4 & 466 & 471 & 470 & That's where the birds were singing in the tree \\
\midrule
\multirow{4}{*}{Qwen3-0.6B 16K (25, 180)} & baseline & 7 & 0 & 173 & 0 & 0 & And the heart is filled with love \\
 & suppress-only & 1 & 0 & 179 & 0 & 0 & soles, and the joy of the occasion \\
 & inject-only & 11 & 1 & 168 & 3 & 0 & I am honored to be here today \\
 & suppress+inject & 3 & 2 & 175 & 1 & 1 & I am honored to be here \\
\bottomrule
\end{tabular}
\caption{Composition-horizon results, steering at the line-3 newline, pooled over prompts (four on the Gemma cells, three on Llama, one on Qwen3) and three seeds. Inject words: around (Gemma 426K), can (Gemma 2.5M), that (Llama), myself (Qwen3 16K). ``Inserted'' counts lines containing the injected word anywhere, and how many of those open with it; the rest occur mid-line except for 9 lines, across all cells and conditions, where it occurs only as the final word. Greedy lines are from the first prompt at seed 1 and match the earlier single-prompt runs. Under emission-adjacent steering the same intervention reaches a median $P(\text{inject})$ of 0.88, 0.74, 0.999 and 0.009 at the sweep peak, against $6.7\times10^{-5}$, $1.1\times10^{-6}$, $6.4\times10^{-4}$ and $2.8\times10^{-6}$ at the newline.}
\label{tab:horizon}
\end{table*}

\begin{table}[h]
\centering\footnotesize
\begin{tabular}{@{}R{4.1cm}rrrr@{}}
\toprule
Criterion for an inject-group ending & base & sup & inj & s+i \\
\midrule
Registered metric (no control) & 21 & 17 & 31 & 38 \\
Final word is not the injected word & 17 & 13 & 14 & 23 \\
Injected word not before the final word & 20 & 17 & 4 & 4 \\
\quad and no word thrice, 4+ distinct & 20 & 17 & 0 & 0 \\
\quad and no word twice, 5+ distinct & 15 & 14 & 0 & 0 \\
Injected word absent from the line & 17 & 13 & 0 & 0 \\
\bottomrule
\end{tabular}
\caption{Redirect recount on Gemma 426K, the only cell with counts above 4 under any criterion (720 lines per condition). Fisher exact, baseline against suppress-and-inject: $p=0.03$, $0.42$, $1.4\times10^{-3}$, $1.7\times10^{-6}$, $5.7\times10^{-5}$, $1.4\times10^{-5}$ down the rows. The nominal rise under the registered metric reverses into a significant fall as soon as lines containing the inserted word are excluded. On the other three cells every criterion gives 0 or 1 under steering. Exploratory, not pre-registered.}
\label{tab:sensitivity}
\end{table}

\paragraph{Pair-level re-analysis.}
Table~\ref{tab:pairs} summarizes the committed suppress-and-inject position sweeps of \citet{pliprs2026}: four prompts per model, every alternative rhyme group discovered for the CLT, fixed $s{=}10$, one sweep per (prompt, inject feature) pair. Because the recorded baseline field of those files disagrees with the flat in-sweep level by a roughly constant factor on Gemma, we use each sweep's own floor (the median over positions) as the reference; a pair counts as detectable when its best position exceeds ten times that floor.

\begin{table}[h]
\centering\footnotesize
\begin{tabular}{@{}lrrr@{}}
\toprule
 & G 426K & L 524K & G 2.5M \\
\midrule
Pairs & 136 & 44 & 264 \\
Argmax at final token & 95 & 36 & 164 \\
Detectable ($>10\times$ floor) & 82 & 33 & 132 \\
\quad at final & 82 & 33 & 132 \\
Behavioral ($P\geq0.1$) & 9 & 3 & 16 \\
\quad at final & 9 & 3 & 16 \\
Argmax at a newline & 7 & 5 & 17 \\
\quad their max ratio & $1.2\times$ & $2.6\times$ & $1.3\times$ \\
Newline/floor, median & $1.01\times$ & $1.15\times$ & $1.05\times$ \\
Newline/floor, max & $1.40\times$ & $3.26\times$ & $1.31\times$ \\
\bottomrule
\end{tabular}
\caption{Pair-level localization and newline inertness at $s{=}10$ (G: Gemma 2 2B, L: Llama 3.2 1B). Every pair with a detectable effect spikes at the final token; no pair's best newline position exceeds $3.3\times$ its floor.}
\label{tab:pairs}
\end{table}

\begin{figure*}[t]
\centering
\includegraphics[width=1\textwidth]{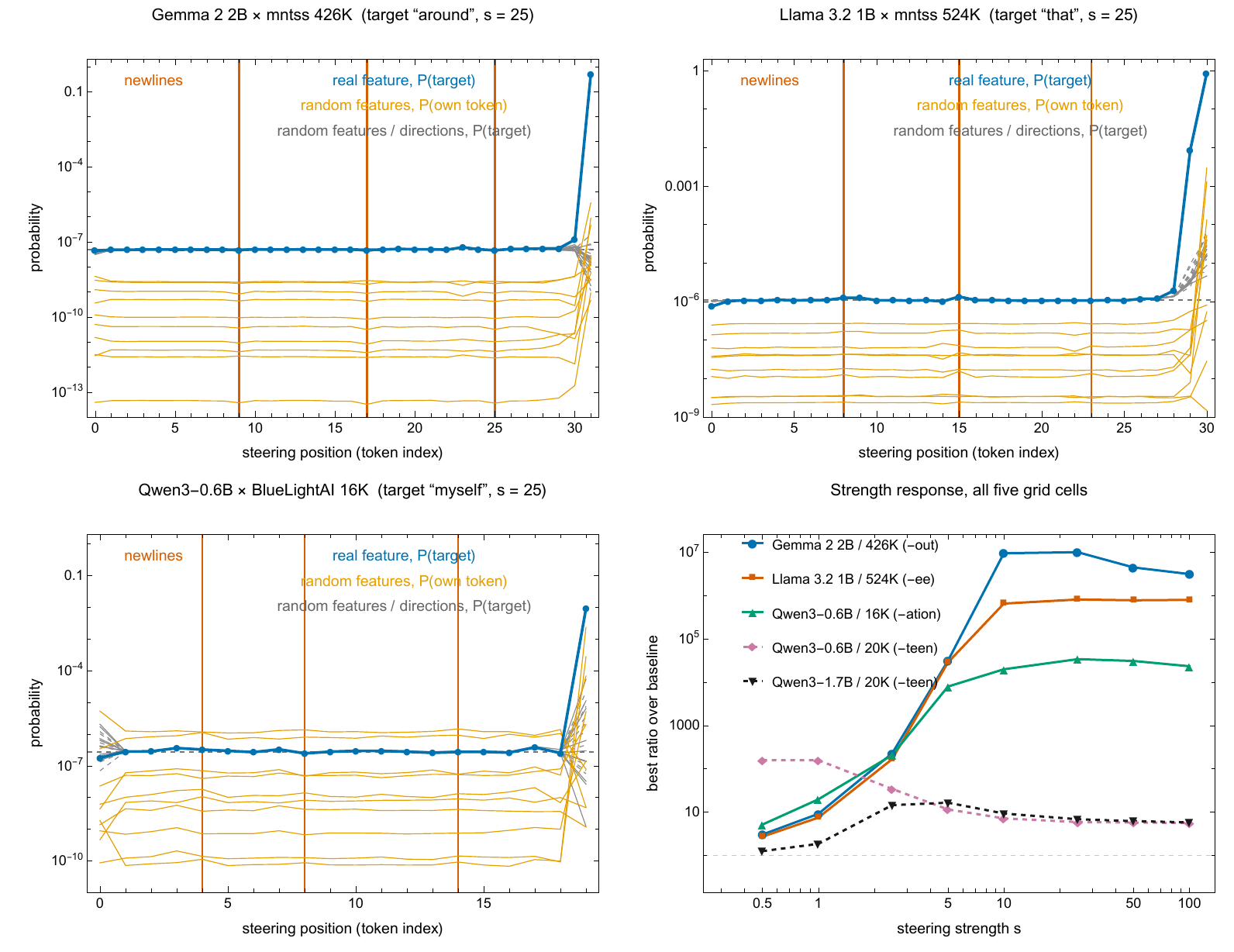}
\caption{Top row and bottom left: per-position random controls on the three cells above the logit-only tier, $s{=}25$. Blue: the real inject feature's $P(\text{target})$. Gray: $P(\text{target})$ under 10 layer-matched random CLT features (solid) and 10 norm-matched random directions (dashed). Orange: each random feature's probability of its \emph{own} top decoder token. Vertical orange lines mark newline positions; the dashed horizontal line is the target's baseline. Random write-directions spike their own token at the final position, but to probabilities at most $3\times10^{-3}$; nothing moves at any newline. Bottom right: best ratio over baseline as a function of steering strength (log-log) for the five grid cells of Table~\ref{tab:cells} with a strength profile; no single strength is near-optimal across cells (\S\ref{sec:ablations}).}
\label{fig:random}
\end{figure*}

\paragraph{Random controls (registered).}
Predictions registered before running: (i) P(target) stays flat under every random draw; (ii) a random feature may spike its own top decoder token at the final position (not a failure: it generalizes the decoder-direction regime); (iii) the random-weight model shows no target spike, with seed-unstable behavior. Outcomes are reported in \S\ref{sec:generic} and \S\ref{sec:ablations}, and per position in Figure~\ref{fig:random}; beyond those numbers, (iii) was confirmed across 3 random-init and 3 weight-shuffle seeds, and the pre-registered $10\times$ \emph{ratio} bound of (i) is recorded as ill-posed for common-token targets and amended to the absolute bound, as disclosed there.

\paragraph{Localization null model.}
With $n_i=[32,31,20,22,22,20,20]$: all seven cells within their last two positions, $\prod_i(2/n_i)=3.3\times10^{-8}$; all seven exactly at the final token, $\prod_i(1/n_i)=2.6\times10^{-10}$; the observed six-of-seven-at-final pattern, $4.2\times10^{-8}$ (exact Poisson-binomial with rates $1/n_i$; a simple binomial at the mean rate agrees within 10\%). Per-prompt breadth: 4/4 prompts at the final token in each reference cell, joint null $7.2\times10^{-13}$.

\section{Reproducibility Lessons from the Implementation}
\label{app:lessons}

\paragraph{Encoder hook.}
Our reference detection code read the CLT encoder from the post-MLP residual. MLP-output reconstruction settles which residual the mntss CLTs were trained on: reconstructing the MLP output from the pre-MLP input gives cosine 0.945 at layer 25, while the post-MLP residual reconstructs worse than predicting zero. Under the correct hook, the three Gemma rhyme features previously reported active at the final token (0.98, 0.25, 0.36) read exactly 0.000. Decoder-side validation (which the steering results rest on, and which is unaffected) did not catch this; only a reconstruction check did.

\paragraph{Tokenizer and stack drift.}
A tokenizer BOS assumption silently measured the wrong sub-token on Qwen3 (`` myself'' resolved to ``self''), and inference-stack drift moved a headline number from 0.777 to 0.687 on identical weights. Both argue for token-identity audits, committed artifacts, and pinned stacks.

\paragraph{Baseline fields.}
The pair-level sweep files of \citet{pliprs2026} record a baseline probability that disagrees with the flat in-sweep level by a roughly constant factor on Gemma (about $50\times$). All ratios in Table~\ref{tab:pairs} are therefore computed against each sweep's own floor; the seven-cell grids of Table~\ref{tab:cells}, produced by the later harness, record a baseline that matches the in-sweep floor.

\end{document}